\documentclass[5p,final,authoryear,twocolumn]{elsarticle}

\usepackage{amssymb}
\usepackage{graphicx}
\usepackage{amsmath}
\usepackage{subfigure}
\usepackage[latin1]{inputenc}
\usepackage{color}
\usepackage{multirow}
\usepackage[hypcap=false]{caption}
\usepackage{rotating}
\usepackage{url}

\DeclareMathOperator*{\argmax}{arg\,max}

\DeclareRobustCommand\onedot{\futurelet\@let@token\@onedot}
\def\@onedot{\ifx\@let@token.\else.\null\fi\xspace}

\definecolor{gold}{rgb}{0.85,.66,0}

\graphicspath{{pics/}{figs/}{figs/segOverlays/}{../Miccai/figs/}{../Miccai/}{figs/figuresRev/}{figs/figures/}}

\usepackage{booktabs}
\journal{NeuroImage}

\begin{document}
\begin{frontmatter}

%

\title{Bayesian QuickNAT: Model Uncertainty in Deep Whole-Brain Segmentation for Structure-wise Quality Control }

\author{Abhijit Guha Roy$^{a,b}$ \corref{cor1}} 
\cortext[cor1]{{Corresponding Author. Address: KJP, LMU, Waltherstr. 23, 80337 M\"unchen, Germany; Email: abhi4ssj@gmail.com}}

\author{Sailesh Conjeti$^{c}$, Nassir Navab$^{b,d}$, Christian Wachinger$^{a}$}

\address{$^{a}$Artificial Intelligence in Medical Imaging (AI-Med), Department of Child and Adolescent Psychiatry, LMU M\"{u}nchen, Germany \\
$^{b}$Computer Aided Medical Procedures, Department of Informatics, Technical University of Munich, Germany \\
$^{c}$German Center for Neurodegenerative Diseases (DZNE), Bonn, Germany \\
$^{d}$Computer Aided Medical Procedures, Johns Hopkins University, Baltimore, USA.
}

\begin{abstract}
We introduce Bayesian QuickNAT for the automated quality control of whole-brain segmentation on MRI T1 scans. 
Next to the Bayesian fully convolutional neural network, we also present inherent measures of segmentation uncertainty that allow for  quality control per brain structure.  
For estimating model uncertainty, we follow a Bayesian approach, wherein, Monte Carlo (MC) samples from the posterior distribution are generated by keeping the dropout layers active at test time. 
Entropy over the MC samples provides a voxel-wise model uncertainty map, whereas expectation over the MC predictions provides the final segmentation.
Next to voxel-wise uncertainty, we introduce four metrics to quantify structure-wise uncertainty in segmentation for quality control.
We report experiments on four out-of-sample datasets comprising of diverse age range, pathology and imaging artifacts. 
The proposed structure-wise uncertainty metrics are highly correlated with the Dice score estimated with manual annotation and therefore present an inherent measure of segmentation quality. 
In particular, the intersection over union over all the MC samples is a suitable proxy for the Dice score. 
In addition to quality control at scan-level, we propose to incorporate the structure-wise uncertainty as a measure of confidence to do reliable group analysis on large data repositories. 
We envisage that the introduced uncertainty metrics would help assess the fidelity of automated deep learning based segmentation methods for large-scale population studies, as they enable automated quality control and group analyses in processing large data repositories.
\end{abstract}


\end{frontmatter}

\section{Introduction}
\label{sec:intro}

Automated brain segmentation is a basic tool for processing magnetic resonance imaging (MRI) and provides imaging biomarkers of neuroanatomy like volume, thickness, and shape. 
Despite efforts to deliver robust segmentation results across scans from different age groups, diseases, field strengths, and manufacturers, inaccuracies in the segmentation outcome are inevitable~\citep{Mindcontrol}.
A fundamental limitation of existing methods for whole-brain segmentation is that they do not estimate segmentation quality. 
Hence, manual quality control (QC) is advised before continuing with the analysis, but it has several shortcomings: (i) time consuming, (ii) subject to intra- and inter-rater variability, (iii) binary (pass/fail), and (iv) global for the entire scan. 
In particular when operating on large datasets, manual QC is very time consuming so that cohort-level summary statistics on biomarkers have, for instance, been used for identifying outliers~\citep{sabuncu2016morphometricity}. 
A shortcoming of such heuristics is that they operate decoupled from the actual image and segmentation procedure. 

Bayesian approaches for image segmentation are an alternative because they do not only provide the mode (i.e., the most likely segmentation) but also the posterior distribution of the segmentation. Most of such Bayesian approaches use point estimates in the inference, whereas marginalizing over parameters has only been proposed in combination with Markov Chain Monte Carlo sampling~\citep{iglesias2013improved}  or the Laplace approximation~\citep{wachinger2015contour}. 
Although sampling-based approaches incorporate fewer assumptions, they are computationally intense, especially when used in conjunction with atlas-based segmentation, and thus, have only been used for segmenting substructures but not the whole-brain~\citep{iglesias2013improved}.

Fully convolutional neural networks (F-CNNs) have become the tool of choice for semantic segmentation in  computer vision~\citep{segnet,long2015fully} and medical imaging~\citep{ronneberger2015u}. 
In prior work, we introduced QuickNAT~\citep{roy2017error, roy2018quicknat}, an F-CNN for whole-brain segmentation of MRI T1 scans that has  not  only outperformed existing atlas-based approaches, but also accomplished the segmentation orders of magnitude faster. 
QuickNAT is also much faster than DeepNAT, a previous patch-based approach for brain segmentation with neural networks~\citep{wachinger2018deepnat}.
Although F-CNNs provide high accuracy, they are often poorly calibrated and fail to estimate a confidence margin with the output~\citep{guo2017calibration}. The predictive probability at the end of the network, i.e., the output of the softmax layer, does not capture the model uncertainty~\citep{gal2016dropout}.

Recent progress in Bayesian deep learning utilized the concept of Monte Carlo (MC) sampling via dropout to approximate samples from the posterior distribution~\citep{gal2016dropout}. Dropout has originally been proposed to prevent overfitting during training~\citep{srivastava2014dropout}. Dropout at test time approximates sampling from a Bernoulli distribution over network weights. As dropout layers do not have learnable parameters, adding them to the network does not increase model complexity or decrease performance. 
Thanks to fast inference with CNNs, multiple MC samples can be generated to reliably approximate the posterior distribution in acceptable time. 
MC dropout for estimating uncertainty in deep learning was originally proposed for classification~\citep{gal2016dropout} and later applied to semantic segmentation  with F-CNNs in computer vision~\citep{kendall2015bayesian}, providing a pixel-wise model uncertainty estimate.

In this article, we propose to inherently measure the quality of whole-brain segmentation with a Bayesian extension of QuickNAT. 
For this purpose, we add dropout layers to the QuickNAT architecture, which enables highly efficient Monte Carlo sampling. 
Thus, for a given input brain scan, multiple possible segmentations are generated by MC sampling. Next to estimating voxel-wise segmentation uncertainty, we propose four metrics for quantifying the segmentation uncertainty for each brain structure. 
We show that these metrics are highly correlated with the segmentation accuracy (Dice score) and can therefore be used to predict segmentation accuracy in absence of reference manual annotation. 
Finally, we propose to effectively use the uncertainty estimates as quality control measures in large-scale group analysis to estimate reliable effect sizes. 

The automated QC proposed in this article offers advantages with regards to manual QC. 
Most importantly, it does not require manual interactions so that an objective measure of quality control is available at the same time with the segmentation, particularly important for processing large neuroimaging repositories. 
Furthermore, we obtain a continuous measure of segmentation quality, which may be a more faithful representation than dichotomizing into pass and fail. 
Finally, the segmentation quality is estimated for each brain structure, instead of a global assessment for the entire brain in manual QC, which better reflects variation in segmentation quality within a scan.


The main contributions of the work are as follows: 
\begin{enumerate}
    \item First approach for whole-brain segmentation with inherent quality estimation
    \item Monte Carlo dropout for uncertainty estimation in brain segmentation with F-CNN
    \item Four metrics to quantify structure-wise uncertainty in contrast to voxel-wise uncertainty  
    \item Comprehensive experiments on four unseen datasets (variation in quality, scanner, pathology) to substantiate the high correlation of structure-wise uncertainty with Dice score
    \item Integration of segmentation uncertainty in group analysis for estimating more reliable effect sizes.
\end{enumerate}

While end-to-end learning approaches achieve high segmentation accuracy, the `black box' nature of complex neural networks may impede their wider adoption in clinical application.
The lack of transparency of such models makes it difficult to trust the outcome. 
In addition, the performance of learning-based approaches is closely tied to the scans used during training. 
If scans are presented to the network during testing that are very different to those that it has seen during training, a lower segmentation accuracy is to be expected. 
With the uncertainty measures proposed in this work, we address these points by also estimating a quality or confidence measure of the segmentation. 
This will allow to identify scans with low segmentation accuracy, potentially due to low image quality or variation from the training set. 
While the contributions in this work do not increase the segmentation accuracy, we believe that assigning a meaningful confidence estimate will be as important for its practical use. 


\section{Prior Art}
Prior work exists in medical image computing for evaluating segmentation performance in absence of manual annotation.
In one of the earliest work, the common agreement strategy (STAPLE) was used to evaluate classifier performance for the task of segmenting brain scans into WM, GM and CSF~\citep{bouix2007evaluating}.
In another approach, the output segmentation map was used, from which features were extracted to train a separate regressor for predicting the Dice score~\citep{kohlberger2012evaluating}.
More recent work proposed the reverse classification accuracy (RCA), whose pipeline involves training a separate classifier on the segmentation output of the method to evaluate, serving as pseudo ground truth~\citep{valindria2017reverse}. Similar to previous approaches, it also tries to estimate Dice score. The idea of RCA was  extended for segmentation quality control in large-scale cardiac MRI scans~\citep{robinson2017}.

In contrast to the approaches detailed above, our approach provides a quality measure or prediction confidence that is \emph{inherently} computed (i.e. derived from the same model, in contrast to using a separate model for estimating quality) within the segmentation framework, derived from model uncertainty. Thus, it does not require to train a second, independent second classifier for evaluation, which itself might be subject to prediction errors. 
An earlier version of this work was presented at a conference~\citep{roy2018inherent} and has here been extended with methodological improvements and  more experimental evaluation.
To the best of our knowledge, this is the first work to provide an uncertainty measure for each structure in whole-brain segmentation and its downstream application in group analysis for reliable estimation. 

\begin{figure}[h]
\centering
\includegraphics[width=0.45\textwidth]{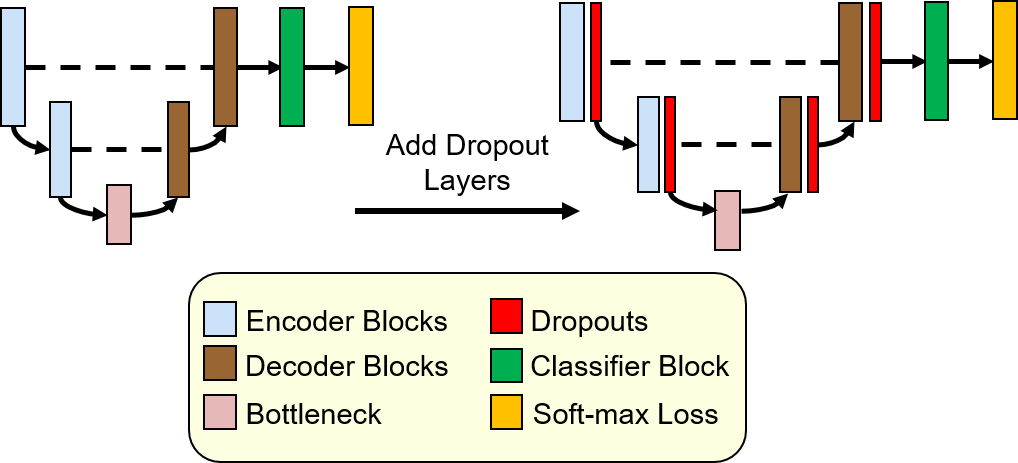}
\caption{Illustration of addition of droupout layers after every encoder and decoder blocks of an F-CNN model to generate Monte Carlo samples of segmentation}
\label{fig:add_dropout}
\end{figure}

\section{Method}
\label{sec:methodology}
We propose a fully convolutional neural network that produces next to the segmentation also an estimate of  the confidence or quality of the segmentation for each brain structure. To this end, we use a Bayesian approach detailed in the following sections.


\subsection{Background on Bayesian Inference}


Given a set of training scans $\mathbf{I} = \{  \tilde{I_1}, \cdots, \tilde{I}_m \}$ with its corresponding manual segmentations $\mathbf{S} = \{  \tilde{S}_1, \cdots, \tilde{S}_m \}$, we aim at learning a probabilistic function $F_{seg}:I \rightarrow S$. This function generates the most likely segmentation $S^\star$ given a test scan $I^\star$. The probability of the predicted segmentation is
\begin{equation}
p(S^\star | I^\star, \mathbf{I}, \mathbf{S}) = \int p(S^\star | I^\star, \mathbf{W}) p(\mathbf{W} | \mathbf{I}, \mathbf{S}) d \mathbf{W},
\label{eqn:integral}
\end{equation}

\noindent
where $\mathbf{W}$ are the weight parameters of the function $F_{seg}(\cdot)$. 
The posterior distribution over weights in Eq.~(\ref{eqn:integral}) is generally intractable, where we use variational inference to approximate it.  
Thus, a variational distribution over network's weights $q(\mathbf{W})$ is learned by minimizing the Kullback-Leibler divergence $KL(q(\mathbf{W})||p(\mathbf{W} | \mathbf{I}, \mathbf{S}))$, yielding the approximate predictive distribution
\begin{equation}
q(S^\star | I^\star, \mathbf{I}, \mathbf{S}) = \int p(S^\star | I^\star, \mathbf{W}) q(\mathbf{W}) d \mathbf{W}.
\label{eqn:approxintegral}
\end{equation}

In Bayesian neural networks, the stochastic weights $\mathbf{W}$ are composed of $L$ layers $\mathbf{W} = (\mathbf{W}_i)_{i=1}^L$. 
The variational distribution $q(\mathbf{W}_i)$ for layer $i$ is sampled as
\begin{align}
\mathbf{W}_i &= \mathbf{M}_i \cdot \mathrm{diag}([z_{i,j}]_{j=1}^{K_i}), \label{eqn:dropout} \\
z_{i,j} &\sim \mathrm{Bernoulli}(p_i), i=1,\dots,L, j=1,\dots,K_{i-1}.  \nonumber
\end{align}

\noindent
Here $z_{i,j}$ are Bernoulli distributed random variables with  probabilities $p_i$, and $\mathbf{M}_i$ are variational parameters to be optimized. The diag$(\cdot)$ operator maps vectors to diagonal matrices whose diagonals are the elements of the vectors. Also, $K_i$ represents the number of nodes in the $i^{th}$ layer.

The integral in Eq.~(\ref{eqn:approxintegral}) is estimated by summing over Monte-Carlo samples drawn from $\mathbf{W} \sim q(\mathbf{W})$. Note that sampling from $q(\mathbf{W}_i)$ can be approximated by performing dropout on layer $i$ in a network whose weights are $(\mathbf{M}_i)_{i=1}^L$~\citep{gal2016dropout}. The binary variable $z_{i,j} = 0$ corresponds to unit $j$ in layer $i-1$ being dropped out as an input to the $i^{th}$ layer. Each sample of $\mathbf{W}$ provides a different segmentation for the same input image. The mean of all the segmentations provides the final segmentation, whereas the variance among segmentations provides model uncertainty for the prediction.

\subsection{QuickNAT architecture}
As the base architecture, we use our recently proposed QuickNAT~\citep{roy2018quicknat}. QuickNAT  consists of three 2D F-CNN models, segmenting an input scan slice-wise along coronal, axial and sagittal axes. This is followed by a view aggregation stage where the three generated segmentations are combined to provide a final segmentation. Each 2D F-CNN model has an encoder-decoder based architecture, four encoder blocks and four decoder blocks separated by a bottleneck block. Dense connections are added within each encoder and decoder block to promote feature re-usability and promote learning of better representations~\citep{huang2017densely}. Skip connections exist between each encoder and decoder block similar to U-Net~\citep{ronneberger2015u}. 
The network is trained by optimizing the combined loss function of weighted Logistic loss and Dice loss. 
Median frequency balancing is employed to compensate for class imbalance~\citep{roy2018quicknat}.

\subsection{Bayesian QuickNAT}
We use dropout layers~\citep{srivastava2014dropout} to introduce stochasticity during inference with the QuickNAT architecture. 
A dropout mask generated from a Bernoulli distribution $z_{i,j}$ generates a probabilistic weight $\mathbf{W_i}$, see Eq.~(\ref{eqn:dropout}),  with random neuron connectivity similar to a Bayesian neural network~\citep{gal2016dropout}.
For Bayesian QuickNAT, we insert dropout layers after every encoder and decoder block with a dropout rate $r$, as illustrated in Fig.~\ref{fig:add_dropout}. 
Dropout is commonly used during training of neural networks to prevent over-fitting, but deactivated during testing. 
Here, we keep dropout active in the testing phase and generate multiple segmentations from the posterior distribution of the model. 
%
%
%
To this end, the input scan $I$ is feed-forwarded $N$ times through QuickNAT, each time with a different and random dropout mask. This process simulates the sampling from a space of sub-models with different connectivity among the neurons. This MC sampling of the models generates $N$ samples of predicted probability maps $\{P_1, \cdots P_N\}$, from which hard segmentation maps $\{S_1, \cdots S_N\}$ can be inferred by the `$\argmax$' operator across the channels $c$. This approximates the process of variational inference as in Bayesian neural networks~\citep{gal2016dropout}. The final segmentation $S$ is estimated by computing the average over all the MC probability maps, followed by a `$\arg\max$' operator as
\begin{equation}
    S = \argmax_c \frac{1}{N} \sum_{i=1}^N P_i.
    \label{eqn:finalseg}
\end{equation}

\noindent
The probability map $P_i$  consists of $c$ channels $\{ P_i^1 \cdots P_i^c \}$, representing probability maps for each individual class, which includes the addressed brain structures and background.

\begin{figure*}[t]
\centering
\includegraphics[width=0.95\textwidth]{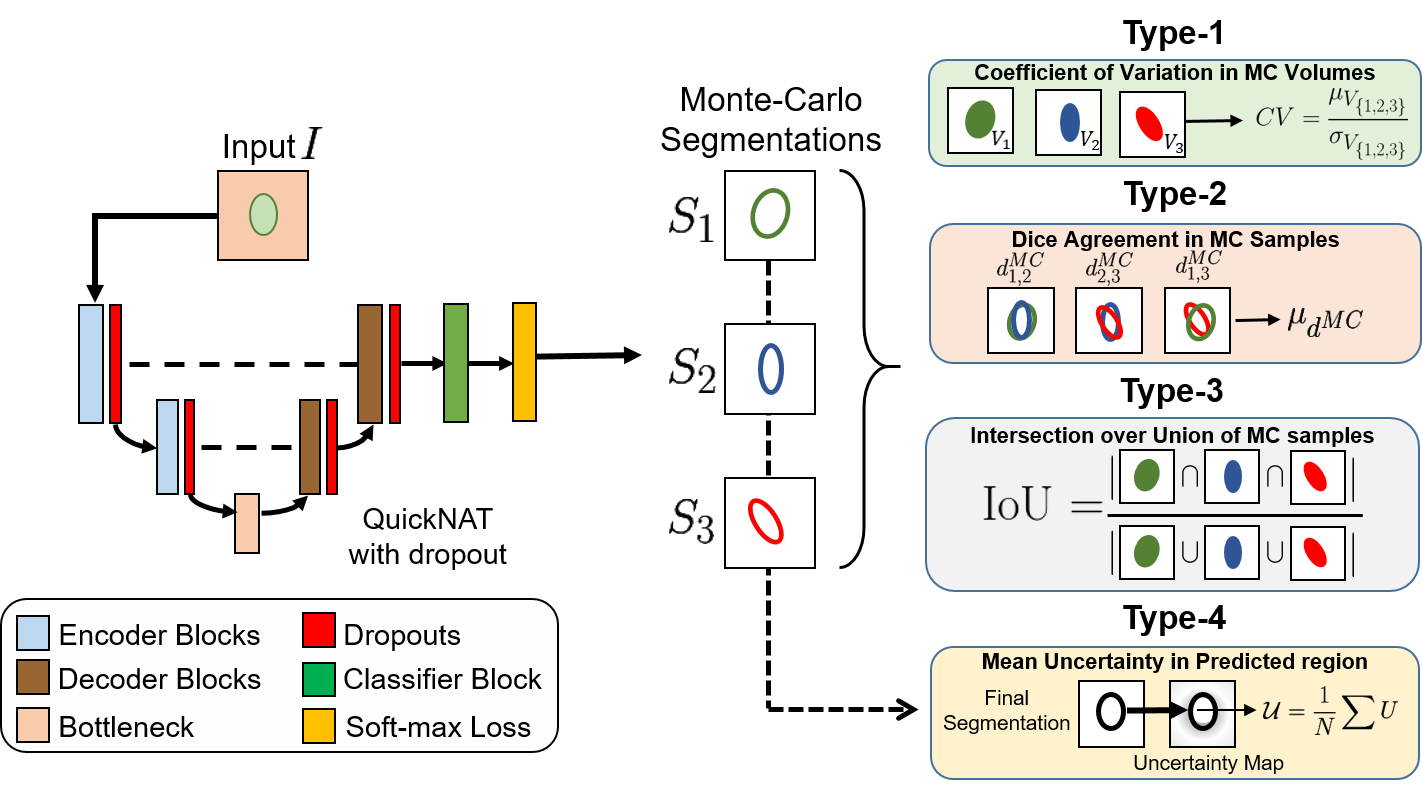}
\caption{A single input scan results in different Monte Carlo (MC) segmentations ($S_1,S_2,S_3$) based on different dropouts in the fully ConvNet.
The samples are used to estimate  three variants of structure-wise uncertainty. 
The final segmentation $S$ is the average of the MC samples as shown in Eq.~\ref{eqn:finalseg}, used in the third variant.}
\label{fig:graphAbs}
\end{figure*}

\subsection{Uncertainty Measures}

\subsubsection{Voxel-wise Uncertainty}
The model uncertainty $U_s$ for a given voxel $\mathbf{x}$, for a specific structure $s$ is estimated as entropy over all $N$ MC probability maps~$\{ P_1^s, \cdots, P_N^s \}$

\begin{equation}
U_s(\mathbf{x}) = - \sum_{i=1}^N P_i^s(\mathbf{x}) \cdot \log(P_i^s(\mathbf{x})).
\label{eq:unc}
\end{equation}

\noindent
The global voxel-wise  uncertainty is the sum over all structures, $U = \sum_s U_s$.  
Voxels with low uncertainty (i.e. low entropy) receive the same predictions, with different random neurons being dropped out from the network. 
An intuitive explanation for this is that the network is highly confident about the decision and that the result does not change much when the neuron connectivity is partially changed by using dropouts. 
In contrast, the prediction confidence is low, if predictions change a lot with altering neuron connectivity.

\subsubsection{Structure-wise Uncertainty}
As most quantitative measures extracted from segmentation maps (e.g., Hippocampus volume) relate to specific brain structures, it is helpful to have an uncertainty measure corresponding to each brain structure, rather than each voxel. Here, we propose four different metrics for computing structure-wise uncertainty from MC segmentations, illustrated in  Fig.~\ref{fig:graphAbs} for  $N=3$ MC samples. 

\noindent
\textbf{Type-1: }
We measure the variation of the volume across the MC samples. As volume estimates are commonly used for neuroanatomical analysis, this type of uncertainty provides a confidence margin with the estimate. We compute the coefficient of variation,
\begin{equation}
CV_s = \frac{\sigma_s}{\mu_s},
\end{equation}
\noindent
with mean~$\mu_s$ and standard deviation~$\sigma_s$ of structure $s$  for MC volume estimates. 
Note that this estimate is agnostic to the size of the structure.

\noindent
\textbf{Type-2: }We use the overlap between samples as a measure of uncertainty. 
To this end, we compute the average Dice score over all possible pairs of $N$ MC samples,
\begin{equation}
d_s^{MC} = E \left[ \{ Dice((S_i=s), (S_j=s)) \}_{i \neq j} \right].
\end{equation}
\noindent
This measures the agreement in area overlap between all the MC samples in a pair-wise fashion. 

\noindent
\textbf{Type-3: }We use the intersection over overlap (IoU$_s$) metric, over all the $N$ MC samples for a specific structure $s$ as measure of its uncertainty. The value of IoU$_s$ is constraint between $[0,1]$ and it is computed as  
\begin{equation}
\mathrm{IoU}_s = \frac{|(S_1=s)\cap(S_2=s)\cap \cdots \cap (S_N=s)|}{|(S_1=s)\cup(S_2=s)\cup \cdots \cup (S_N=s)|}.
\end{equation}

\noindent
\textbf{Type-4: }We define the uncertainty for a structure $s$ as mean global voxel-wise uncertainty over the voxels which were labeled as $s$,  
\begin{equation}
\mathcal{U}_s = E \left[  \{ U(\mathbf{x}) \}_{\mathbf{x}\in \{ S=s \}}  \right].
\end{equation}

It must be noted that $d_s^{MC}$ and IoU$_s$ are directly related to segmentation accuracy, while $\mathcal{U}_s$ and $CV_s$ are inversely related to accuracy. Also, it is worth mentioning that computing voxel-wise uncertainty maps requires all $N$ segmentation \emph{probability maps} $P_i$ (each one having a size around 2 GB), which can be computationally  demanding. In contrast, our proposed metrics (except $\mathcal{U}_s$) use \emph{label maps} $S_i$ (size around 200 KB), which are much smaller in size and can be computed faster.

\subsection{Segmentation Uncertainty in Group Analyses}
Commonly, image segmentation is only a means to an end, where image-derived measures are used in follow-up statistical analyses. 
We are interested in propagating the uncertainty from the segmentation  to the follow-up analyses. 
The rationale is that segmentations with high uncertainty potentially corresponds to scans with poor quality whose inclusion would confound the true effect sizes and limit the statistical significance of observed group differences.
We demonstrate the integration of uncertainty for generalized linear models (GLMs) in the following, but it can also be generalized to other statistical models. 
GLMs are frequently used in neuroimaging studies for identifying significant associations between image measures and variables of interest. 
For instance, in numerous group analyses studies Hippocampus volume was shown to be an important imaging biomarker with significant associations to Alzheimer's disease. 


In solving the  regression model, each equation, i.e., each subject,  has equal importance in the optimization routine (i.e. $\omega_i=1, \forall i$). 
In contrast, we propose to integrate the structure-wise uncertainty in the analysis. 
This is achieved by solving a weighted linear regression model with an unique weight $\omega_i\geq0$ for subject~$i$,
\begin{equation}
\hat{\boldsymbol{\beta}} = \arg \min \sum_i \omega_i (V_i - \mathbf{X}_i \boldsymbol{\beta}^\top)^2,
\label{eq:weightedAnalysis}
\end{equation}
with design matrix $\mathbf{X}$, vector of coefficients $\boldsymbol{\beta}$, and normalized brain structure volume  $V_i$ (normalized by intra cranial volume). 
We use the proposed structure-wise uncertainties ($CV_s$, $d_s^{MC}$ and IoU$_s$) and set the weight as,
\begin{equation}
\omega_i=\frac{1}{CV_s}, \quad  \quad \omega_i = \frac{1}{1-d_s^{MC}}\quad \mathrm{or} \quad \omega_i = \text{IoU}_s.
\label{eqn:weights}
\end{equation}
\noindent

Including  weights in  the regression increases its robustness as scans with reliable segmentation are emphasized. 
Setting all weights to a constant results in standard regression. 
In our experiments, we use 
\begin{equation}
\mathbf{X}_i = [1, A_i, S_i, D_i]  \quad \quad    \boldsymbol{\beta} = [\beta_0, \beta_A, \beta_{S}, \beta_{D}]
\end{equation}
with age $A_i$, sex $S_i$ and diagnosis $D_i$ for subject $i$. 
Of particular interest is the regression coefficient $\beta_{D}$, which estimates the effect of diagnosis on the volume of a brain structure. 

\section{Experimental Setup}

\subsection{Architecture and Training Procedure}
We set the dropout rate to $r=0.2$ (other values of $r$ decreased the segmentation performance compared to not using droupouts) and produce $N=15$ MC samples ($<$ 2 minutes), after which performance saturates (shown in Sec.~\ref{sec:numMC}).
For training the neural network with limited data, we use the pre-training strategy with auxiliary labels proposed earlier~\citep{roy2017error}. 
To this end, we pre-train the network on $581$ volumes of the IXI dataset\footnote{http://brain-development.org/ixi-dataset/} with segmentations produced by  FreeSurfer~\citep{fischl2002whole}  and subsequently fine-tune on $15$ of the 30 manually annotated volumes from the Multi-Atlas Labelling Challenge (MALC) dataset~\citep{landman2012miccai}. The remaining $15$ volumes were used for testing. The split is consistent to challenge instructions. This trained model is used for all our experiments. In this work, we segment $33$ brain structures (listed in the appendix).

\subsection{Test Datasets}
\label{sec:dataset}
We test of four datasets, where three of the datasets have not be seen during training. 

\begin{enumerate}
\item \textbf{MALC-15}: 15 of the 30 volumes from the MALC dataset that were not used for training are used for testing. MALC is a subset of the OASIS repository~\citep{marcus2007open}.
\item \textbf{ADNI-29}: The dataset consists of 29 scans from Alzheimer's Disease Neuroimaging Initiative (ADNI) database (adni.loni.usc.edu), with a balanced distribution of Alzheimer's Disease (AD) and control subjects, and scans acquired with 1.5T and 3T scanners. The objective is to observe uncertainty changes due to variability in scanner and pathologies. The ADNI was launched in 2003 as a public-private partnership, led by Principal Investigator Michael W. Weiner, MD. For up-to-date information, see www.adni-info.org.
\item \textbf{CANDI-13}: The dataset consists of 13 brain scans of children (age 5-15) with psychiatric disorders, part of the CANDI dataset~\citep{kennedy2012candishare}. The objective is to observe changes in uncertainty for data with age range not included in training. 
\item \textbf{IBSR-18}: The dataset consist of 18 scans publicly available at \url{https://www.nitrc.org/projects/ibsr}. The objective is to see the sensitivity of uncertainty with low resolution and poor contrast scans.
\end{enumerate}

Note that the training set (MALC) did not contain scans with AD or scans from children. 
Manual segmentations for MALC, ADNI-29, and CANDI-13 were provided by Neuromorphometrics, Inc.\footnote{http://Neuromorphometrics.com/}

\section{Experimental Results and Discussion}
\label{sec:exp}

\subsection{Number of MC Samples}
\label{sec:numMC}
First, we examine the choice of number of MC samples ($N$) needed for our task. 
This choice is mainly dependent on two  factors: (i) the segmentation accuracy by averaging all the MC predictions needs to be similar to the segmentation accuracy not using dropouts at test time, and (ii) the estimated uncertainty map needs to be stable, i.e., addition of more MC samples should not effect the computed entropy values.
We use the CANDI-13 dataset for this experiment as it represents an out of sample dataset, i.e., data not used in training the model. It therefore provides a realistic test case on unseen data. We performed experiments with $N=\{ 3,6,9,12,15,18 \}$.


The mean global Dice scores for different values of $N$ are reported in Tab.~\ref{tab:mc_acc}. We observe that the Dice score remains more or less constant as $N$ increases from $3$ to $18$, which is very close to the Dice performance with no dropouts at test time. 
This is in contrast to prior work that reported a performance increase with more MC samples~\citep{kendall2015bayesian}. 
A potential reason for this is that the QuickNAT framework aggregates segmentations across the three principal axes (coronal, axial and sagittal)~\citep{roy2018quicknat}. Hence, $N$ MC samples actually represents aggregating $3 \cdot N$ segmentations in our framework. Furthermore, the view aggregation step compensates from the slight decrease in segmentation performance due to dropout at test time. 

\begin{table}[t]
\centering
\small
\caption{Mean Dice scores on CANDI dataset with different number of MC samples and without dropout.}
  \begin{tabular}{|c|c|}
    \hline
     \#MC samples ($N$) & Mean Dice score \\
    \hline
    3 &  $0.801\pm 0.035$  \\ 
    6 &  $0.803\pm0.033$ \\ 
    9 &  $0.803\pm0.036$ \\ 
    12 & $0.804\pm0.034$ \\
    15 & $\mathbf{0.806}\pm0.037$ \\ 
    18 & $\mathbf{0.807}\pm0.034$ \\ \hline
    No Dropout & $\mathbf{0.806}\pm0.035$ \\ \hline
  \end{tabular}
  \label{tab:mc_acc}
\end{table}

Next, we investigate the number of MC samples needed to reliably estimate the model uncertainty. The voxel-wise uncertainty can be considered stable if the estimated entropy values do not change substantially with larger $N$. 
Let the uncertainty maps for $i$ and $j$ MC samples be $U_i$ and $U_j$, respectively. 
We estimate the mean absolute difference between them, $E[ | U_i - U_j | ]$  to quantify the stability. We report this value for different consecutive transitions $i \rightarrow j$ of MC samples in Tab.~\ref{tab:mc_unc}. We observe that the transition $15 \rightarrow 18$ yields a small difference, indicating a stable estimation of the uncertainty maps.

\begin{table}[t]
\centering
\caption{Mean absolute change in voxel-wise Entropy Map $E[abs(U_i-U_j)]$, when entropy estimated from using $i$ MC samples ($U_i$) to using $j$ MC samples ($U_j$). }
  \begin{tabular}{|c|c|}
    \hline
     Transitions ($i \rightarrow j$) & $E[ |U_i - U_j| ] \times 10^{-3}$ \\
    \hline
    $3 \rightarrow 6$ & $0.7827$  \\ 
    $6 \rightarrow 9$ & $0.5135$ \\ 
    $9 \rightarrow 12$ & $0.3539$ \\ 
    $12 \rightarrow 15$ & $0.2421$ \\
    $15 \rightarrow 18$ & $\mathbf{0.0925}$ \\ \hline
  \end{tabular}
  \label{tab:mc_unc}
\end{table}

It is worth mentioning that as $N$ increases, not only does the segmentation time per scan increase, but also the required computational resources and complexity. 
This is due to the fact that all the $N$ intermediate 3D segmentation probability maps (4D tensors) need to be loaded in the RAM for estimating the voxel-wise uncertainty map. 
We set $N=15$ for all the following experiments, which provides high segmentation accuracy and reliable uncertainty estimates, while keeping the computational complexity within acceptable margins.

\begin{table*}[ht]
\centering
\caption{Results on 4 different datasets with global Dice scores (DS) and correlation of Dice scores with 4 types of uncertainty.}
  \begin{tabular}{|p{0.80in}| c|p{0.40in}|p{0.40in}|p{0.40in}|p{0.40in}|p{0.40in}|}
    \hline
     Datasets & Mean Dice score  & \multicolumn{4}{c|}{Corr($\cdot$, DS$_s$)} & Mean \\
      &  & $\mathcal{U}_s$ & $CV_s$ & $d_s^{MC}$ & IoU$_s$ & IoU$_s$  \\
    \hline
    \textbf{MALC-15} & $\mathbf{0.88}\pm0.02$ & $-0.85$ & $-0.81$ & $0.86$ & $\mathbf{0.88}$ & $0.88$  \\ 
    \textbf{ADNI-29} & $0.83\pm0.02$  & $-0.72$ & $-0.71$ & $\mathbf{0.78}$ & $\mathbf{0.78}$ & $0.85$   \\ 
    \textbf{CANDI-13} & $0.81\pm0.03$ & $-0.84$ & $-0.86$ & $0.90$ & $\mathbf{0.91}$ & $0.82$  \\ 
    \textbf{IBSR-18} & $0.81\pm0.02$ & $-0.76$ & $-0.76$ & $0.80$ & $\mathbf{0.84}$ & $0.83$\\ \hline
  \end{tabular}
  \label{tab:res}
\end{table*}

\subsection{Uncertainty based quality control across different datasets}
\label{sec:dice_corr}
\begin{figure*}
\centering
\includegraphics[width=\textwidth]{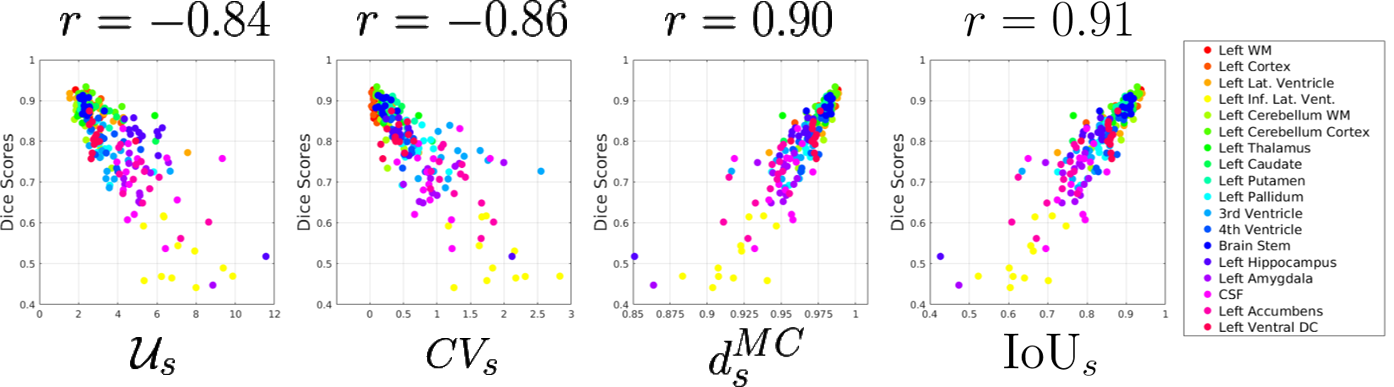}
\caption{Scatter plot of four types of proposed uncertainty and Dice scores  on CANDI-13 dataset (one dot per scan and structure), with their corresponding correlation coefficient ($r$). For clarity, structures only on the left hemisphere are shown. }
\label{fig:scatterplot}
\end{figure*}

\begin{figure*}[h]
\centering
\includegraphics[width=0.75\textwidth]{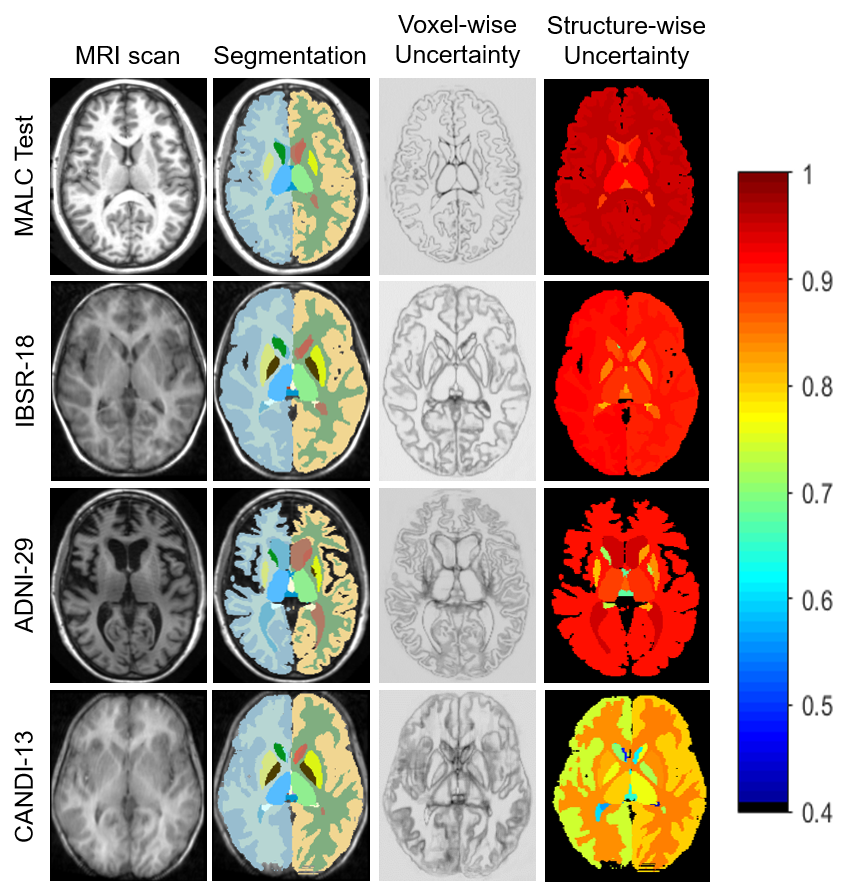}
\caption{Results of 4 different cases, one from each dataset, corresponding to the worst Dice score. The MRI scan, segmentation, voxel-wise uncertainty and structure-wise uncertainty (IoU$_s$) are presented. The color coding of IoU$_s$ heat map between $[0.4,1]$ is shown to the right, with darker shades of red indicating high reliability in segmentation.}
\label{fig:results}
\end{figure*}

In this section, we conduct experiments to explore the ability of the proposed structure-wise uncertainty metrics in predicting the segmentation quality across different seen and un-seen datasets.
Towards this end, we compute the correlation coefficient between the four uncertainty metrics and the Dice scores to quantify its efficacy in providing quality control. 
We report the mean Dice score, the correlation coefficients and mean IoU in Table~\ref{tab:res} for all four datasets described in Sec.~\ref{sec:dataset}.
Firstly, we observed that the segmentation Dice score is the highest on MALC dataset ($88\%$), while the performance drops by $5-7\%$ Dice points for other datasets (ADNI, CANDI and IBSR). The reason for this is that part of the MALC dataset was used for training, whereas the other datasets are un-seen scans resembling more realistic scenarios with training and testing scans coming from different datasets.
This decrease in Dice score is accompanied by decrease in mean IoU (i.e. increase in structure-wise uncertainty). 
We also observe that all the correlation values with the four metrics for all datasets are within acceptable margins ($0.71-0.91\%$). IoU  has the highest correlation  across all  four datasets. 
Next to reporting correlations, we show the scatter plots for the four uncertainty measures with respect to actual Dice score on CANDI-13 dataset in Fig.~\ref{fig:scatterplot}. In the scatter plots, we represent one dot per structure per scan, with unique colors for each of the classes. For the sake of clarity, structures from the left hemisphere of the brain are only displayed. 
We note that $d_s^{MC}$ and IoU$_s$ show compact point clouds, whereas $\mathcal{U}_s$ is more dispersed indicating lower correlation. 
It must be noted that each of the three unseen datasets has unique characteristics, which are not present in the training MALC scans. IBSR consists of scans with low resolution and thick slices. ADNI contains subjects exhibiting neurodegenerative pathologies, whereas training was done on healthy subjects. CANDI consists of children scans, whereas none of the training subjects was from that particular age range. So, we believe our experiments cover a wide variability of out of sample data (resolution, pathology, age range), which the model might encounter in a more uncontrolled setting. This is shown in Fig.~\ref{fig:results} and explained in detail in Sec.~\ref{sec:quality}.


\begin{table}[ht]
\centering
\caption{Comparison between Dice scores (DS) and IoU$_s$. Correlation and mean absolute error (MAE) between Dice score and IoU, together with accuracy as identifying segmentations as bad, medium, and good.}
  \begin{tabular}{|p{2cm}|c|c|c|}
    \hline
     Dataset & Corr(IoU, DS) & MAE & Accuracy  \\
    \hline
    \textbf{MALC-15} & $0.88$ & $0.02$ & $0.88$   \\ 
    \textbf{ADNI-29} & $0.78$ & $0.07$ & $0.83$ \\ 
    \textbf{CANDI-13} & $0.91$ & $0.04$ & $0.84$\\ 
    \textbf{IBSR-18} & $0.84$ & $0.06$ & $0.80$\\ \hline
  \end{tabular}
  \label{tab:iou}
\end{table}

\subsection{$\mathrm{IoU}$ as a proxy for the Dice score}
The Dice coefficient is the most widely used metric for evaluating segmentation accuracy and provides an intuitive `goodness' measure to the user. This has motivated earlier works to directly regress the Dice score for segmentation quality control~\citep{kohlberger2012evaluating,valindria2017reverse}. 
Our approach is different because we provide inherent measures of  uncertainty of the segmentation model. 
While we have demonstrated that our measures are highly correlated to  Dice scores (Sec.~\ref{sec:dice_corr}), the actual structure-wise uncertainty values may be challenging to interpret because it is not immediately clear which values indicate a good or bad segmentation. 
When looking at the scatterplot in Fig.~\ref{fig:scatterplot}, we see that the uncertainty measures on the x-axis and the Dice score on the y-axis are in different ranges, with the only exception of IoU. 
Indeed, the values of IoU closely resembles the Dice score and we will demonstrate in the following paragraph that it is a suitable proxy for the Dice score. 

We estimated the mean absolute error (MAE) between IoU$_s$ and Dice score and reported the results in Table~\ref{tab:iou}. Also, similar to \cite{valindria2017reverse}, we define three categories, i.e., Dice range $[0.0,0.6)$ as `bad', $[0.6,0.8)$ as `medium' and $[0.8,1.0]$ as `good'. We categorize the segmentations with actual Dice score and IoU$_s$, and report the per-class classification accuracy in Table \ref{tab:iou}. 
MAE varies between $2-7\%$, while accuracy between $80-88\%$ as reported in Table \ref{tab:iou}. 
All the similarity metrics (Correlation, MAE and 3-class classification accuracy) between IoU$_s$ and Dice score have values very similar or better to the ones reported in \citep{valindria2017reverse} over 4 different datasets. 
This is remarkable because \cite{valindria2017reverse} trained a model to dedicatedly  predict the Dice score, while we are simply computing the intersection over overlap of the MC samples without any supervision. 

\begin{figure*}[t]
\centering
\includegraphics[width=\textwidth]{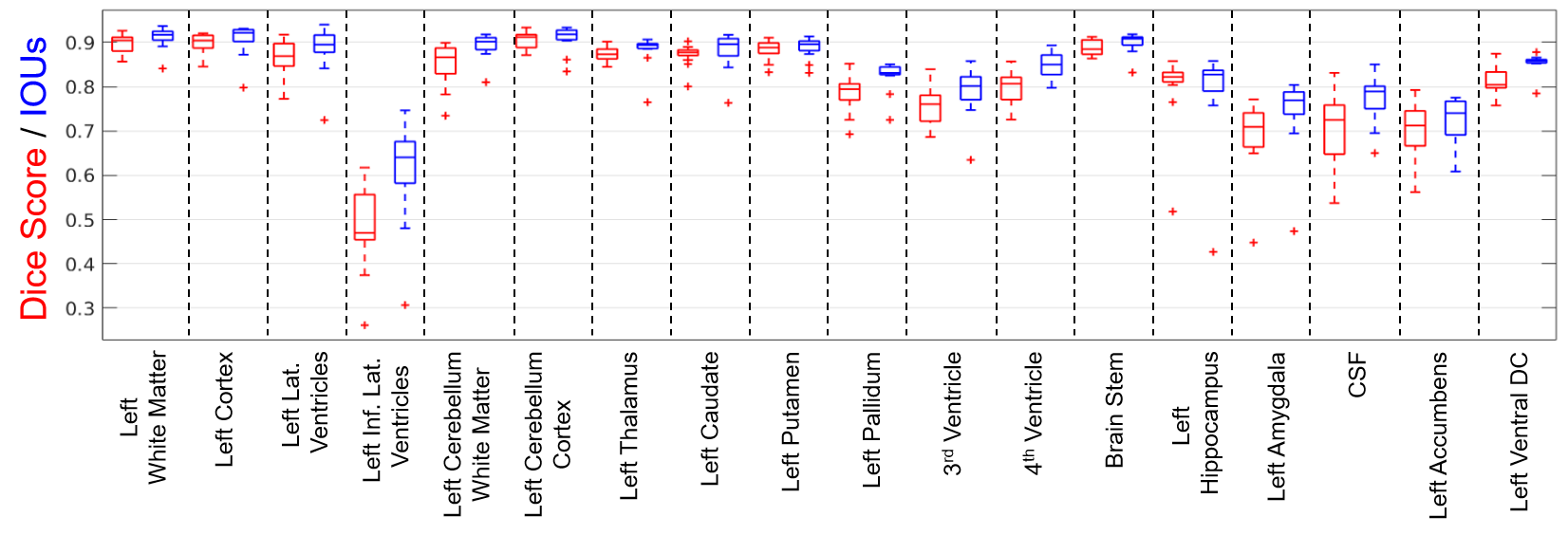}
\caption{Boxplot of Dice score (in \textcolor{red}{red}) and IoU$_s$ (in \textcolor{blue}{blue}) per structure on CANDI-13 dataset. Only structures on the left hemisphere of the brain is shown for clarity. Center-lines indicate the median, boxes extend to the 25th and 75th percentiles, and the whiskers reach to the most extreme values not considered outliers (indicated by \textcolor{red}{red} crosses).}
\label{fig:boxplotplot}
\end{figure*}

We also presented a structure-wise analysis to investigate similarity between Dice score and IoU$_s$ in Fig.~\ref{fig:boxplotplot}. 
Again, only structures on the left hemisphere of the brain are shown for clarity. In the boxplot, we observe that for most of structures IoU$_s$ is very close to actual Dice score. The worst similarity is observed for the  inferior lateral ventricles, where there is about 15\% difference between the two metrics. A potential reason could the small size of the structure.
With all these experiments, we substantiate the fact that IoU$_s$ can be effectively used as a proxy for actual Dice score, without any reference manual annotations.

\subsection{Sensitivity of Uncertainty to scan Quality}
MRI scans of poor quality can lead to a degradation of the segmentation performance. Such poor quality scans can  occur due to various reasons like noise, motion artifacts, and poor contrast. Model uncertainty is expected to be sensitive to the scan quality and should increase whenever segmentation accuracy decreases due to poor data quality. In this section, we investigate whether this property holds for our proposed model. Towards this end, we performed an experiment where we artificially degraded the quality of the input brain MRI scan with Rician noise. Here we use the MALC test dataset for evaluation purposes. We corrupt the scans with dB levels $\{ 3,5,7,9 \}$ and reported the mean global Dice score and mean IoU$_s$ at each noise level in Tab.~\ref{tab:noise}. We observe that the mean Dice score reduces as the dB level of the added Rician noise increases, whereas mean IoU$_s$ also decreases (indicating an increase in uncertainty). This confirms our hypothesis than our model is sensitive to scan quality. We also observe that mean IoU$_s$ falls at a faster rate than mean Dice score, indicating that uncertainty is more sensitive to noise than segmentation accuracy. It must be noted that in all our experiments with real scans, we did not encounter any scenario where segmentation failed (Dice score $< 0.5$). The experiment with Rician noise with $dB=9$ resembles an artificially induced failure case. 

\begin{table}[t]
\centering
\caption{Effect of different Rician noise levels on mean Dice scores, mean IoU$_s$, mean absolute error (MAE) and accuracy of identifying segmentations as bad, medium, and good.}
  \begin{tabular}{|p{1.35cm}|p{1.8cm}|p{1cm}|c|c|}
    \hline
     Noise Levels & Mean Dice score & Mean IoU$_s$ & MAE & Accuracy  \\
    \hline
    No Noise & $\mathbf{0.88\pm0.02}$ & $\mathbf{0.88}$ & $\mathbf{0.02}$ & $\mathbf{0.88}$   \\ 
    dB = 3 & $0.87\pm0.02$ & $0.83$ & $0.05$ & $0.83$ \\ 
    dB = 5 & $0.85\pm0.03$ & $0.78$ & $0.10$ & $0.75$ \\ 
    dB = 7 & $0.69\pm0.18$ & $0.58$ & $0.26$ & $0.72$ \\
    dB = 9 & $0.37\pm0.25$ & $0.22$ & $0.25$ & $0.70$ \\ \hline
  \end{tabular}
  \label{tab:noise}
\end{table}

\subsection{Qualitative Analysis}
\label{sec:quality}
We present  qualitative results of Bayesian QuickNAT in Fig.~\ref{fig:results}. From left to right,  the input MRI scan, its corresponding segmentation, voxel-wise uncertainty map and structure-wise uncertainty (IoU$_s$) heat map are illustrated.
The scale of the heat map replicates the Dice score $[0,1]$, where red corresponds to $1$, indicating higher reliability in segmentation.
Each row shows an example from the four different datasets, where we selected the scan with the worst segmentation accuracy for each dataset.
The first row shows results on a test sample from the MALC dataset, where segmentation is overall of high quality. This is reflected by the thin lines in the voxel-wise uncertainty (anatomical boundaries) and redness in the structure-wise uncertainty heat map. 
Since the same dataset was used for training, we obtain high segmentation accuracy on MALC.
The second row presents the scan with worst performance on the IBSR-18 dataset. Careful inspection of the MRI scan shows poor contrast with prominent ringing artifacts. The mean Dice score of the scan is $0.79$, which is $3\%$ below the mean score for the overall dataset. 
An increase in voxel-wise uncertainty can be observed visually by the thickening of the lines along anatomical boundaries (in contrast to MALC). The structure-wise uncertainty maps shows lighter shades of red in some sub-cortical structures, indicating a lesser reliable segmentation, in comparison to MALC. 
The third row presents the scan with worst performance in ADNI-29, which belongs to a subject of age 95 with severe AD pathology. Prominent atrophy in cortex along with enlarged ventricles can be visually observed in the scan. In addition to the pathology, ringing artifacts at the top of the scan can be observed. The mean Dice score  is $0.78$, which is $5\%$ below the mean Dice score for the dataset.
Its IoU$_s$ heat map shows higher uncertainty in some subcortical structures with brighter shades, whereas the reliability of cortex and lateral ventricles segmentation is  good. It must be noted that training scans did not consist of any subjects with AD, and this example illustrates the performance of our framework for un-seen pathology.
The last row presents the MRI scan with the worst performance on CANDI-13. The mean Dice score of the scan is $0.73$, which is $8\%$ below the mean Dice performance of the dataset. This scan can be considered as an outlier in the dataset. The scan belongs to a subject of age 5 with strong motion artifacts together and poor contrast. Scans of such age range and such poor quality were not used in the training pipeline, which explain the degradation of the segmentation performance. Its voxel-wise uncertainty is higher in comparison to others, with some prominent dark highly uncertain patches in subcortical regions. 
The heat map shows the lowest confidence for this scan, in comparison to other results. The cortical regions show shades of yellow, whereas some sub-cortical structures show shades of blue, which is towards the lower end of the reliability scale.

\subsection{Uncertainty for Group Analysis}
In the following section, we integrate structure-wise uncertainty in regression models for robust group analyses.

\subsubsection{Group analysis on ADNI-29}
ADNI-29 is a small subset of the ADNI dataset with 15 control and 14 Alzheimer's patients. 
We perform a group analysis  as per Eq.~(\ref{eq:weightedAnalysis}) with age, sex, and diagnosis as independent variables and the volume of a brain structure as independent variable. 
Since we have manual annotations for ADNI-29, we can compute the actual volumes  and accordingly estimate the ground truth regression coefficients. 
Table~\ref{tab:linearReg} reports the regression coefficients for diagnosis  $\beta_D$ for  twelve brain structures. 
The coefficients are estimated based on manual annotations, segmentation with FreeSurfer and with Bayesian QuickNAT. 
Further, we use the uncertainty-based weighting on the volume measures from Bayesian QuickNAT. 
Weighting was done using three of the proposed structure-wise uncertainty as presented in Eq.~(\ref{eqn:weights}). 
Our hypothesis  is that weighting will result in regression coefficients $\beta_D$ that are numerically equal or closer to the estimates from the manual annotation than those without weighting.  
We observe that out of the selected 12 structures, more reliable estimation of $\beta_D$ is achieved with weighting and five structures using $d_s^{MC}$ based weighting. Also for all structures, any weighting resulted in $\beta_D$ estimation, which is closer to its actual value, thus substantiating our hypothesis. 
These results demonstrate that integrating segmentation quality in the statistical analysis leads to more reliable estimates. 

\begin{table*}[t]
\centering
\caption{Estimates of regression coefficient for diagnosis $\beta_D$ in group analyses on ADNI-29 consisting of healthy controls and AD patients. Results are reported for manual annotations and segmentations with FreeSurfer and Bayesian QuickNAT. For the volume measures with Bayesian QuickNAT, we also report uncertainty-based weighting with $CV_s$, $d_s^{MC}$ and IoU$_s$.}
  \begin{tabular}{|c|c|c|c|c|c|c|}
      \hline
     \textbf{Structures} & Manual & FreeSurfer & QuickNAT & \multicolumn{3}{c|}{Bayesian QuickNAT} \\ 
     & Annotations & &  &  $CV_s$ & $d_s^{MC}$ & IoU$_s$ \\
    \hline
\textbf{White Matter}	&	1.129	& 0.788 &	0.779	&	0.778	&	0.779	&	\textbf{0.799}	\\
\textbf{Cortex}	&	-0.202	& -0.406 &	-0.156	&	-0.158	&	-\textbf{0.177}	&	-0.146	\\
\textbf{Lateral ventricle}	&	-0.368	& -0.392 &	-\textbf{0.372}	&	-0.376	&	-0.423	&	-0.405	\\
\textbf{Caudate}	&	-0.111	& -0.026 &	-0.047	&	-0.088	&	-\textbf{0.131}	&	-0.067	\\
\textbf{Putamen}	& 0.109	& 0.225 &	0.276	&	0.237	&	0.055	&	\textbf{0.130}	\\
\textbf{3rd Ventricle}	&	-0.214	& -0.333 &	-0.353	&	-0.357	&	-0.391	&	-\textbf{0.325}	\\
\textbf{4th Ventricle}	&	-0.022	& -0.055 &	-0.076	&	-0.063	&	-0.019	&	-\textbf{0.022}	\\
\textbf{Hippocampus}	&	1.149	& 0.979 &	1.282	&	1.280	&	1.249	&	\textbf{1.191}	\\
\textbf{Amygdala}	&	1.005	& 0.891 &	1.149	&	1.104	&	\textbf{1.039}	&	0.908	\\
\textbf{Accumbens}	&	0.343	& 0.738 &	0.516	&	0.469	&	\textbf{0.384}	&	0.473	\\
    \hline
\end{tabular}
  \label{tab:linearReg}
\end{table*}


\subsubsection{ABIDE-I}
We perform group analysis on the ABIDE-I dataset~\citep{di2014autism} consisting of $1,112$ scans, with $573$ normal subjects and $539$ subjects with autism. The dataset is collected from 20 different sites with a high variability in scan quality. 
To factor out changes due to site, we added site as a covariate in Eq.~\ref{eq:weightedAnalysis}. 
We report $\beta_{D}$  with corresponding p-values for the volume of brain structures that have recently been associated to autism in a large ENIGMA study~\citep{van2017cortical}.
We compare uncertainty weighted regression (weighted by $CV_s$, $d_s^{MC}$ and IoU$_s$) to normal regression in Table~\ref{tab:groupAnalysis}.
Strikingly, uncertainty weighted regression results in significant associations to autism, identical to \citep{van2017cortical}, whereas normal regression is only significant for amygdala.

Standard approaches for group analysis on large cohorts involves detection of outlier volume estimates and removing the corresponding subjects from the regression process. This sometimes also requires a manual inspection of the segmentation quality. In contrast to these approaches, we propose to use all the scans and associated a continuous weight for all, providing their relative importance is estimating the regression coefficients without the need for any outlier detection or manual inspection. 

\begin{table*}[ht]
\small
\centering
\caption{Results of group analyses on ABIDE dataset with autism pathologies, with and without using uncertainty. $\star$ indicates statistical significance in reported p-values.}
  \begin{tabular}{|p{1in}|c|c|c|c|c|c|c|c|}
      \hline
     \textbf{Autism} & \multicolumn{2}{c|}{Normal Regression} & \multicolumn{2}{c|}{$CV_s$} & \multicolumn{2}{c|}{$d_s^{MC}$} &
      \multicolumn{2}{c|}{IoU$_s$} \\ 

      \textbf{Biomarkers}& $\beta_{D}$ & $p_{D}$ & $\beta_{D}$ & $p_{D}$  & $\beta_{D}$ & $p_{D}$  & $\beta_{D}$ & $p_{D}$ \\
    \hline
    \textbf{Amygdala} & $-0.14$ & $0.0140^\star$ & $-0.32$ & $0.0001^\star$ & $-0.27$ & $0.0001^\star$ & $-0.19$ & $0.0012^\star$  \\ 
    \textbf{Lat. Ventricles} & $-0.01$ & $0.8110$ & $-0.38$ & $0.0089^\star$ & $-0.19$ & $0.0843$ & $-0.07$ & $0.0442^\star$\\ 
    \textbf{Pallidum} & $-0.07$ & $0.2480$ & $-0.40$ & $0.0051^\star$ & $-0.28$ & $0.0165^\star$ & $-0.25$ & $0.0322^\star$ \\
    \textbf{Putamen} & $-0.07$ & $0.2186$ & $-0.43$ & $0.0035^\star$ & $-0.39$ & $0.0057^\star$ & $-0.37$ & $0.0059^\star$\\
    \textbf{Accumbens} & $-0.08$ & $0.1494$ & $-0.21$ & $0.0013^\star$ & $-0.17$ & $0.0031^\star$ & $-0.12$ & $0.0421^\star$ \\ \hline
  \end{tabular}
  \label{tab:groupAnalysis}
\end{table*}

\subsection{General Discussion}
We introduced an approach to not only estimate the segmentation but also the uncertainty in the segmentation. 
The uncertainty is directly estimated from the segmentation model. 
Consequently, the uncertainty increases if a test scan is presented to the network that is different to the scans that it has seen during training. 
On the one hand, this holds for individuals that have different demographic characteristics or pathologies. 
On the other hand, this holds for image quality, which is related to the image acquisition process. 
Learning-based approaches can produce staggering segmentation accuracy, but there is strong dependence on the scans used during training. 
Since it will be impossible to have all scans that can potentially occur in practice represented in the training set, uncertainty is a key concept to mark scans with lower segmentation accuracy. 
Uncertainty could therefore be used to decide if scans have to acquired again due to insufficient quality. 
Further, it could be used to guide the inclusion of particular types of scans in training. 

Our experiments have demonstrated that structure-wise uncertainty measures are highly correlated to the Dice score. 
They can therefore be used for automated quality control. 
In particular, the intersection over union of the Monte Carlo samples has the same range as the Dice score and is demonstrated to be highly correlated with Dice in unseen datasets.
Consequently, it can be interpreted as a proxy for the Dice score when manual annotations are not available to compute the actual Dice score. 
This can be beneficial for judging the segmentation quality of single scans. 

For the analysis of groups of images, we then went one step further and integrated uncertainty measures in the follow-up analysis. 
We have demonstrated the impact of such an integration for regression analysis, but the general concept of weighting instances by their uncertainty can be used for many approaches, although it may require some adaptation. 
Such an approach offers particular advantage for the analysis of large repositories, where a manual quality control is very time consuming. 
Our results for the regression models have shown that weighting samples according to the segmentation quality yields estimates that are more similar to those from the manual annotation. 

\section{Conclusion}
\label{sec:conc}
In this article, we introduced  Bayesian QuickNAT, an  F-CNN  for whole brain segmentation with a structure-wise uncertainty estimate. 
Dropout is used at test time to produce multiple Monte Carlo samples of the segmentation, which are used in estimating uncertainty. 
We introduced four different metrics to quantify structure-wise uncertainty. 
We extensively validated on multiple unseen datasets and demonstrate that the proposed metrics have high correlation with segmentation accuracy and provide effective quality control in absence of reference manual annotation. 
The datasets used in the experiments include unseen data from a wide variety with scans from children, with pathologies, with low resolution and with low contrast. 
Strikingly, one of our proposed metrics, intersection over union of MC samples, closely approximates the  Dice score.  
In addition to this, we proposed to integrate the uncertainty metrics as confidence in the observation into group analysis, yielding reliable effect sizes. 
Although, all the experiments are performed on neuroimaging applications, the basic idea is generic and can easily to extended to other segmentation applications. We believe our framework will aid in translating automated frameworks for adoption in large scale neuroimaging studies as it comes with a fail-safe mechanism to indicate the user whenever the system is not sure about a decision for manual intervention.

\section*{Acknowledgement}
We thank SAP SE and the Bavarian State Ministry of Education, Science and the Arts in the framework of the Centre Digitisation.Bavaria (ZD.B) for funding and the NVIDIA corporation for GPU donation.
We thank Neuromorphometrics Inc. for providing manual annotations. 
Data collection and sharing was funded by the Alzheimer's Disease Neuroimaging Initiative (ADNI) (National Institutes of Health Grant U01 AG024904) and DOD ADNI (Department of Defense award number W81XWH-12-2-0012). ADNI is funded by the National Institute on Aging, the National Institute of Biomedical Imaging and Bioengineering, and through generous contributions from the following:
Alzheimer's Association; Alzheimer's Drug Discovery Foundation; Araclon Biotech; BioClinica Inc.; Biogen Idec Inc.; Bristol-Myers Squibb Company; Eisai Inc.; Elan Pharmaceuticals, Inc.; Eli Lilly and Company; EuroImmun; F. Hoffmann-La Roche Ltd and its affiliated company Genentech, Inc.; Fujirebio; GE Healthcare; IXICO Ltd.; Janssen Alzheimer Immunotherapy Research \& Development, LLC ; Johnson \& Johnson Pharmaceutical Research \& Development LLC; Medpace, Inc; Merck \& Co., Inc.; Meso Scale Diagnostics, LLC; NeuroRx Research; Neurotrack Technologies; Novartis Pharmaceuticals Corporation; Pfizer Inc.; Piramal Imaging; Servier; Synarc Inc.; and Takeda Pharmaceutical Company. 
The Canadian Institutes of Health Research is providing funds to support ADNI clinical sites in Canada. Private sector contributions are facilitated by the Foundation for the National Institutes of Health (www.fnih.org). The grantee organization is the Northern California Institute for Research and Education, and the study is coordinated by the Alzheimer's Disease Cooperative Study at the University of California, San Diego. ADNI data are disseminated by the Laboratory for Neuro Imaging at the University of Southern California.

\section*{Appendix}

\textbf{List of Classes: } (1) Left white matter, (2) Left cortex, (3) Left lateral ventricle, (4) Left inferior lateral ventricle, (5) Left cerebellum white matter, (6) Left cerebellum cortex, (7) Left thalamus, (8) Left caudate, (9) Left putamen, (10) Left pallidum, (11) $3^{rd}$ ventricle, (12) $4^{th}$ ventricle, (13) Brain stem, (14) Left hippocampus, (15) Left amygdala, (16) CSF, (17) Left accumbens, (18) Left ventral DC, (19) Right white matter, (20) Right cortex, (21) Right lateral ventricle, (22) Right inferior lateral ventricle, (23) Right cerebellum white matter, (24) Right cerebellum cortex, (25) Right thalamus, (26) Right caudate, (27) Right putamen, (28) Right pallidum, (29) Right hippocampus, (30) Right amygdala, (31) Right accumbens, (32) Left ventral, and (33) Optic Chiasma.

\section{References}

\bibliographystyle{elsarticle-harv}
\bibliography{papers}

\end{document}